\documentclass[letterpaper]{article} 
\usepackage{aaai23}  
\usepackage{times}  
\usepackage{helvet}  
\usepackage{courier}  
\usepackage[hyphens]{url}  
\usepackage{graphicx} 
\usepackage{natbib}  
\usepackage{caption} 
\usepackage{algorithm}
\usepackage{algorithmic}

\usepackage{subcaption}
\usepackage{sectsty}
\sectionfont{\normalsize}
\subsectionfont{\normalsize}
\subsubsectionfont{\normalsize}
\paragraphfont{\normalsize}
\usepackage{makecell}
\usepackage{color}
\usepackage{stfloats}
\usepackage{booktabs}
\usepackage{newtxtext,newtxmath}
\usepackage{tabularx}
\usepackage{multirow}
\usepackage[toc,page]{appendix}

\usepackage{newfloat}
\usepackage{listings}
\DeclareCaptionStyle{ruled}{labelfont=normalfont,labelsep=colon,strut=off} 
\floatstyle{ruled}
\newfloat{listing}{tb}{lst}{}
\floatname{listing}{Listing}
\title{Look Around! A Neighbor Relation Graph Learning Framework \\ for Real Estate Appraisal}
\author{
    Chih-Chia Li, Wei-Yao Wang, \textsuperscript{\rm 1}Wei-Wei Du, Wen-Chih Peng
}
\affiliations{
    National Yang Ming Chiao Tung University, Hsinchu, Taiwan\\
    \{licc.cs09, \textsuperscript{\rm 1}wwdu.cs10\}@nycu.edu.tw, sf1638.cs05@nctu.edu.tw, wcpeng@cs.nycu.edu.tw
}

\begin{document}

\maketitle

\begin{abstract}
Real estate appraisal is a crucial issue for urban applications, which aims to value the properties on the market.
Traditional methods perform appraisal based on the domain knowledge, but suffer from the efforts of hand-crafted design.
Recently, several methods have been developed to automatize the valuation process by taking the property trading transaction into account when estimating the property value.
However, existing methods only consider the real estate itself, ignoring the relation between the properties.
Moreover, naively aggregating the information of neighbors fails to model the relationships between the transactions.
To tackle these limitations, we propose a novel Neighbor \textbf{Re}lation \textbf{Gra}ph Learning Fra\textbf{m}ework (\textbf{ReGram}) by incorporating the relation between target transaction and surrounding neighbors with the attention mechanism.
To model the influence between communities, we integrate the environmental information and the past price of each transaction from other communities.
Moreover, since the target transactions in different regions share some similarities and differences of characteristics, we introduce a dynamic adapter to model the different distributions of the target transactions based on the input-related kernel weights.
Extensive experiments on the real-world dataset with various scenarios demonstrate that ReGram robustly outperforms the state-of-the-art methods.
Furthermore, comprehensive ablation studies were conducted to examine the effectiveness of each component in ReGram. 

\end{abstract}

\section{Introduction}
\label{sec:introduction}

Property technology (proptech) has developed proprietary systems by bringing properties and their owners from offline to online and has stimulated several productive studies for digital marketing, \textit{e.g.}, virtual tours and online appraisal, especially due to the COVID-19 situation \cite{PropTech}.
In this paper, we focus on one of the proptech applications, real estate appraisal, which come to play a vital role, not only for the society but also for the government, as it reduces the burden of triad relationships (\textit{e.g.,} buyers, sellers, and owners) when bargaining for the proper value of a property.
For example, real estate appraisal provides an objective price to prevent buyers from being deliberately sold overpriced properties by sellers.
Furthermore, it also avoids underestimating or overestimating the value of the property, which assists the owners in more reasonably measuring their assets.
On the other hand, the price of real estate can be viewed as one of the development metrics in urban areas; thus real estate appraisal benefits the government in terms of planning urban development.
Therefore, it is essential to develop real estate appraisal algorithms for eliminating unfair trades and boosting economic prosperity.

\begin{figure}
  \centering
  \includegraphics[width=0.80\linewidth]{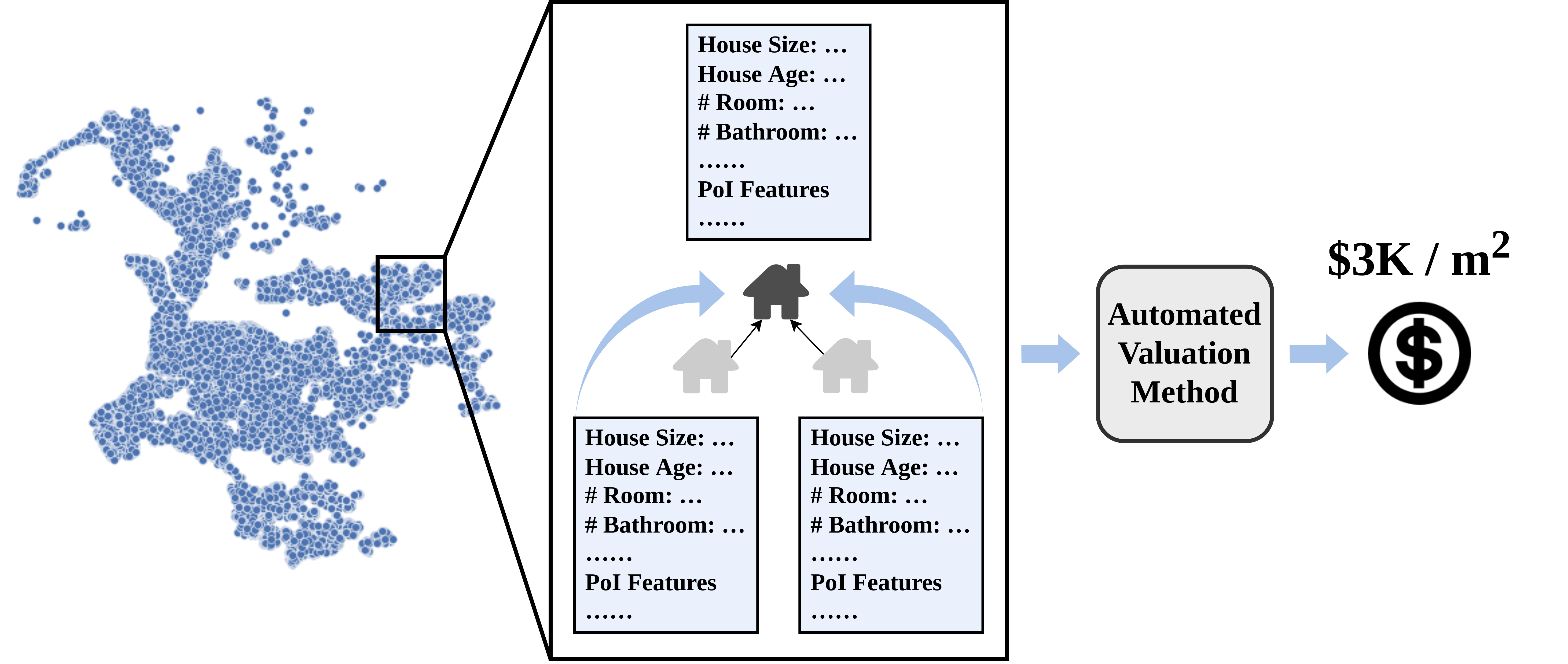}
  \caption{Illustration of a real scenario in Taipei with AVMs for real estate appraisal.}
  \label{fig:1}
\end{figure}

Generally, there are several domains in the real world that are also tasked with modeling information according to such geographic location for analysis, \textit{e.g.}, rental pricing \cite{DBLP:conf/kdd/YeQCWZMYZ18} and traffic forecasting \cite{DBLP:conf/icml/LanMHWYL22}.
A key solution for taking the surrounding information into account is to adopt a graph neural network (GNN) (Fig. \ref{fig:1}), which has drawn significant attention due to its capability of modeling non-Euclidean data like social networks \cite{DBLP:conf/www/SankarLYS21} and financial trading \cite{cheng2022financial,yin2022graph}.
In real estate appraisal, the only existing approach exploits GNN by considering the neighbor information from different perspectives and dividing the final prediction as multitask learning based on the urban district of the target transaction to model value distribution in different regions \cite{DBLP:conf/kdd/0003LZZLD021}.
Despite the above progress, there are three shortcomings in the previous work.
First, there is some irrelevant information of real estate neighbors (\textit{e.g.,} number of rooms), which would not directly affect the price per unit area of other transactions, even if they are close.
Therefore, naively fusing whole features of neighbor transactions will hamper the model performance due to the noise.
Second, the values of the properties are significantly influenced by the geographic location, which can represent the regional economic situation of the target transaction.
For instance, if a property is located in a wealthy area, the corresponding value is more likely to be expensive.
However, existing works ignore that the value of surrounding properties will affect the target transaction, which neglects the impact of the geographic location.
Third, when sharing some similarities and differences of characteristics with the transactions in different regions, the existing works fail to consider that there are few transactions in some regions (\textit{e.g.,} remote areas), which causes underfitting and worsens the performance.
Therefore, we believe that real estate appraisal with modeling neighbor information is still an unsolved but essential problem.

To tackle the above challenges, we present a novel Neighbor \textbf{Re}lation \textbf{Gra}ph Learning Fra\textbf{m}ework (\textbf{ReGram}) for real estate appraisal.
For the first issue, we aggregate the relation between the target transactions and their neighbors by learning the weighted neighbor relationship.
For the second issue, in order to consider the impact of geographic location, we introduce a preliminary appraisal of real estate based on the neighbors' prices, which provides surrounding price information for the target property.
For the third issue, we introduce a dynamic adaptor to consider the discrepancies between each target transaction for modeling the distribution of the real estate value.

In summary, the main contributions of our paper are as follows:
\begin{enumerate}
    \item We propose ReGram, a novel neighbor relation graph learning framework to appraise real estate by leveraging the neighborhood relations and their corresponding prices, which can be applied to other geographic urban applications (\textit{e.g.}, rental pricing and traffic forecasting).
    \item We introduce a preliminary appraisal of the target transaction based on the neighbors' prices to provide surrounding price information.
    Moreover, we further propose a dynamic adaptor to flexibly model the price of target transactions based on different characteristics.
    \item Extensive experiments show that ReGram achieves a state-of-the-art performance on the real-world real estate appraisal dataset with various scenarios.
\end{enumerate}

\section{Preliminaries}
\subsection{Related Work}
\label{sec:related-work}

To estimate the value of properties, traditional approaches conduct the process of valuation based on domain experts \cite{mccluskey1997evaluation,baum2013income}, but lack the ability to appraise automatically.
In recent years, several automated valuation methods (AVMs) have been proposed by adopting machine learning \cite{lin2011predicting,DBLP:journals/eswa/AhnBOK12,DBLP:conf/gecco/AzimluRM21} and deep learning techniques \cite{DBLP:journals/tist/LawPR19,DBLP:conf/icdm/GeWXLZ19} but ignoring spatially proximal real estate \cite{fu2014exploiting}.
Besides, several works have utilized artificial neural networks or big data approach \cite{DBLP:conf/icdm/GeWXLZ19,DBLP:journals/tist/LawPR19,bin2019attention,DBLP:journals/symmetry/LeeKH21} to appraise the real estate.
However, most of the previous works ignored the spatio-temporal dependencies among real estate transactions.
The information of peer-dependency were utilized with k-nearest neighbors by sampling a fixed number of similar transactions, and generating sequences from target transactions and nearby transactions for estimating the price of real estate \cite{bin2019peer}.
Still, the sampling process treated each feature as being equally important, which causes sampling noise from unrelated features.

Recently, MugRep, which is the only existing method exploiting GNN in real estate appraisal, was proposed with a multitask hierarchical graph based framework by constructing the graphs at the transaction level and community level \cite{DBLP:conf/kdd/0003LZZLD021}.
To predict the value of real estate, MugRep learned the prediction weights independently by separating the tasks via urban districts.
However, MugRep hampers the model prediction due to considering irrelevant information by aggregating all features of neighbors directly, and fails to take the neighbor transactions' price into account, which is critical information to estimate the value of the target real estate.
Moreover, they predict the value based on district divisions, which ignores the correlation between each region and is prone to incorrect prediction in regions with insufficient transactions.
Our novel approach, in contrast, considers the relation between the target transactions and their neighbors as a weighted relation representation.
Besides, we integrate the neighbors' prices to take the surrounding appraisal information of the target property into account, and dynamically model the distributions based on the target transactions.

\subsection{Definitions of Real Estate Appraisal}

\textbf{Definition 1: Real Estate Transaction}. Consider a set of property trading records as real estate transactions $\boldsymbol{S}$ in chronological order; a real estate transaction at the $t$-th transaction denotes $\boldsymbol{s_t}=\{s^e_t, s^o_t\} \in \boldsymbol{S}$ where $s^e_t$ is the environment feature and $s^o_t$ is the object feature.

\noindent\textbf{Definition 2: Real Estate Value}. Consider a set of property unit prices as real estate value $\boldsymbol{P}$ in chronological order; a real estate value at the $t$-th transaction denotes $\boldsymbol{p_t}\in \boldsymbol{P}$, where $p_t$ is the value of the real estate transaction.

\noindent\textbf{Definition 3: Target Transaction}. Target transaction $\boldsymbol{s_{t+1}}=\{s^e_{t+1}, s^o_{t+1}\}$ is the real estate transaction that will be appraised.

\noindent\textbf{Definition 4: Environment Feature}. Environment features represent the surrounding environment information of the property. In this paper, we denote the PoI (point of interest) feature, geographic information (longitude, latitude), land usage and house age as environment features.

\noindent\textbf{Definition 5: Object Feature}. Object features represent the information about the property such as the size of the house, the number of rooms in the house, and the floor number of the house, etc.

\noindent\textbf{Definition 6: PoI Feature}. PoI features consist of two types of features. The first is the number of PoI with a Euclidean distance. For example, there are two schools around 5km from the property. The other is the minimum distance to each PoI category. For example, the shortest distance to the hospital or gas station.

\noindent\textbf{Definition 7: Community}. A community is a hyper node containing multiple transactions that are built in the same region at the same time. In this paper, two transactions belong to the same community if and only if they meet the following conditions: (1) The completion dates (month and year) of construction are the same. (2) The distance between two transactions is less than 500m \cite{DBLP:conf/kdd/0003LZZLD021}.

\noindent\textbf{Definition 8: Real Estate Appraisal Problem}. Given a target transaction $\boldsymbol{s_{t+1}}$ with its previous transactions $\boldsymbol{S_{[0..t]}}$ and corresponding unit price $\boldsymbol{P_{[0..t]}}$, the task is to estimate the transaction unit price $\boldsymbol{p_{t+1}}$.

\section{Approach}
\label{sec:model_architecture}

\begin{figure*}
  \centering
  \includegraphics[width=\linewidth]{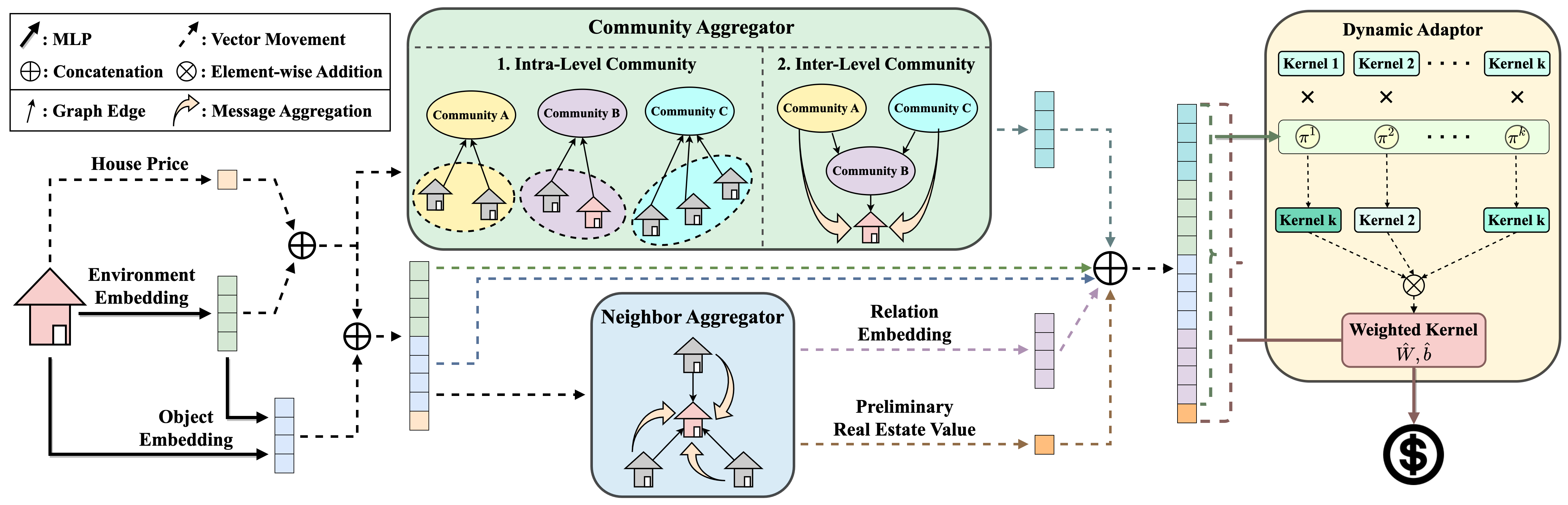}
  \caption{An overview of the proposed ReGram. The community aggregator encodes the transactions with intra-level and inter-level communities. The neighbor aggregator considers the neighbor information of the target transaction to generate the relation representation and preliminary appraisal. The dynamic adaptor generates input-related weights for appraising the real estate.
    }
  \label{fig:ReGram_overview}
\end{figure*}

Figure \ref{fig:ReGram_overview} illustrates an overview of the proposed model.
The graph construction is divided into the transaction-level subgraph, intra-level community and inter-level community for modeling various relationships between transactions and communities.
Specifically, we used the transaction-level subgraph to model the information of surrounding transactions with similar geographic locations and property characteristics.
On the other hand, we constructed the community-level subgraph in order to model the information at the higher level aspect.
Furthermore, the community-level subgraph was divided into the intra-level community and inter-level community to build the connection from transaction to community and the relation between communities, respectively.
After the graph construction, real estate transactions were first embedded by the transaction projection to get transaction representations.
Then we utilized the neighbor aggregator to produce the relation embedding and a preliminary real estate value from surrounding neighbor transactions, and adopted the community aggregator to get the community-level representation.
Afterwards, the dynamic adaptor generated the weighted kernels based on the above contexts and appraised the target transaction.

\subsection{Graph Construction}
\label{sec:graph_construction}

We construct the transaction-level subgraph, intra-level community, and inter-level community similar to \cite{DBLP:conf/kdd/0003LZZLD021}.
However, previous work heavily relies on plentiful and various data (\textit{e.g.}, check-in, and user-trip) for constructing the community; thus, it is hard to generalize to different situations.
Therefore, we introduce the community node using a simple yet effective method by discussing with domain experts, and only used the PoI feature to construct the community-level subgraph, which demonstrates the robustness shown in the experiments.

\noindent\textbf{Transaction-Level Subgraph.}
\label{sec:transaction_level}
To model the relation between transactions, we build the transaction-level subgraph, where each node is a real estate trading transaction.
For each edge, the edge between two nodes needs to satisfy all conditions:

\begin{enumerate}
\item \textbf{Numeric Difference}. The distance between two properties is less than 500m, the trading date difference is no more than 1 year unless they are both in the same month, and the difference of house age is less than 10 years.
\item \textbf{Property Characteristic}. The building type and the property's main purpose (4 and 1622 types, respectively) of the two transactions are required to be the same. For instance, if two properties are both apartment and for commercial-used, then these properties will satisfy this condition.
\item \textbf{Characteristic Indicator}. There are also indicators to characterize the property, including the small house indicator, shop indicator, and first-floor indicator.
The two properties need to have the same indicator results to match this condition.
For example, the two transactions have the same characteristic indicator if both are small and first-floor houses but are not shops.
\end{enumerate}

\noindent\textbf{Intra-Level Community.}
\label{sec:intra_level}
It is expected that surrounding communities will influence the real-estate transaction.
Therefore, the objective of the intra-level community is to model the relation between transaction and community, where each node represents a transaction or a community.
We formulate the community nodes and construct the relation between transaction and community according to Def. 7.
Besides, we filter out the transactions for which the trading time is far from the current target transactions (two months was set in this paper) to avoid considering irrelevant information to the current target time.

\noindent\textbf{Inter-Level Community.}
\label{sec:inter_level}
To model the influence between communities and the target transaction, we designed an inter-level community for modeling the relation between communities, where each node in the inter-level community represents the community.
Specifically, we first averaged the PoI features of transactions in that community to indicate the general living environment in that area.
Then we iterated all community pairs, connecting the communities if the l2 distance of the pair was the top 0.1\% smallest of all the pairs \cite{DBLP:conf/kdd/0003LZZLD021}.

\subsection{Transaction Encoding}
To model the contextualized information of transactions, we first divided the $t$-th transaction features into environment feature $s^e_{t}$ (Def. 4) and object feature $s^o_{t}$ (Def. 5).
The environment embedding $e_{t}$ was obtained by:
\begin{equation}
    e_{t}=W_{e_2}\sigma(W_{e_1}s^e_{t}),
\end{equation}
where $W_{e_1} \in \mathbb{R}^{2d_m \times d_e}$ and $W_{e_2} \in \mathbb{R}^{d_m \times 2d_m}$ are learnable matrices, and $\sigma(\cdot)$ is the Mish activation function \cite{misra2019mish}.

For the object embedding, we first followed the process of  environment embedding to transform the object feature, and we fused with the environment embedding to obtain the object embedding $o_{t}$ for considering the interactions between the property characteristics and the surrounding environment:

\begin{equation}
    o_{t}=W_x\sigma([o'_{t}\oplus e_{t}]); o'_{t}=W_{o_2} \sigma(W_{o_1} s^o_{t}),
\end{equation}
where $W_{o_1} \in \mathbb{R}^{d_m \times d_o}, W_{o_2} \in \mathbb{R}^{d_m \times d_m}$ and $W_x \in \mathbb{R}^{d_m \times 2d_m}$ are learnable matrices, and $\oplus$ is the concatenation operator.

To consider the relation of information and the price of the transaction, we concatenated the environment embedding $e_{t}$, object embedding $o_{t}$, and its price $p_{t}$ as the transaction embedding $x_{t}$:
\begin{equation}
    x_{t}=[e_{t} \oplus o_{t} \oplus p_{t}],
\end{equation}
where $x_{t} \in \mathbb{R}^{d_x}$ and $d_x=2d_m+1$.
It is noted that we set the house price of the target transaction as zero since the target price was the value we wanted to estimate.

\subsection{Neighbor Aggregator}

The neighbor aggregator aims to consider the relations between the target transaction and its neighbors with two steps: relation modeling and relation aggregation.

\noindent\textbf{Relation Modeling.}
We generated the transaction relation $r_{(t+1)t'}$ by projecting the concatenation of the target transaction embedding $x_{t+1}$ and the neighbor transaction embedding $x_{t'}$, where $t'$ indicates the index of the neighbor transaction (Def. 1):

\begin{equation}
    r_{(t+1)t'}=W_r [x_{t+1} \oplus x_{t'}],
\end{equation}
where $W_r \in \mathbb{R}^{d_m \times 2d_x}$ is a learnable kernel weight.


After generating the transaction relation, the attention mechanism is adopted to characterize the importance of each neighbor transaction for the target transaction.
Specifically, the $i$-th head coefficient is computed as follows:
\begin{equation}
    \beta^i_{(t+1)t'}={\rm LeakyReLU}(w_a^i \sigma(r_{(t+1)t'})),
\end{equation}
where $w_a \in \mathbb{R}^{1 \times d_m}$ is a learnable weight.
To extend the potential important neighbors, we computed the attention weight $\alpha^i_{(t+1)t'}$ with softmax temperature $\tau$ as follows:

\begin{equation}
    \alpha^i_{(t+1)t'}=\frac{{\rm exp}(\frac{\beta^i_{(t+1)t'}}{\tau})}{\Sigma_{k \in N_{t+1}}{\rm exp}(\frac{\beta^i_{(t+1)k}}{\tau})},
    \label{eq:neighbor_attention}
\end{equation}
where $N_{t+1}$ is the set of the target transaction's neighbor indexes.

For the multi-head attention, we took the average of all the attention heads to enhance the capability from different perspectives:
\begin{equation}
    \alpha_{(t+1)t'}=\frac{1}{H}\sum \limits_{i=1}^{H} \alpha^i_{(t+1)t'},
\end{equation}
where $H$ is the number of attention heads.

On the other hand, we propose the delta value to reflect the potential price difference between neighbor transaction and target transaction.
The computation of the delta value $d_{(t+1)t'}$ is as follows:

\begin{equation}
    d_{(t+1)t'}=w_d \sigma(r_{(t+1)t'}),
\end{equation}
where $w_d \in \mathbb{R}^{1 \times d_m}$ is a learnable weight.

\noindent\textbf{Relation Aggregation.}
After computing the attention weight and delta value, the relation aggregation produces a preliminary estimated real estate value and the weighted relation embedding from nearby neighbors.

Formally, the neighbor's price is fused with the delta value and aggregate corresponding importance to generate the preliminary value $\tilde{p}_{t+1}$:

\begin{equation}
    \tilde{p}_{t+1}=\sum \limits_{t' \in N_{t+1}} \alpha_{(t+1)t'} (p_{t'}+d_{(t+1)t'}).
\end{equation}

On the other hand, the relation embedding is obtained by aggregating with the attention weights of the neighbors as:

\begin{equation}
    r_{t+1}=\sum \limits_{t' \in N_{t+1}} \alpha_{(t+1)t'} r_{(t+1)t'}.
\end{equation}

\subsection{Community Aggregator}
To further consider the information from a higher level perspective, we designed a community aggregator to integrate the representation from neighbor communities.
Specifically, the environmental representation $\tilde{x}_{t'}$ was computed as: 
\begin{equation}
    \tilde{x}_{t'}=[e_{t'} \oplus p_{t'}],
\end{equation}
where $\tilde{x}_{t'} \in \mathbb{R}^{d_{\tilde{x}}}$ and $d_{\tilde{x}}=d_m+1$.

In the community aggregator, we calculated the attention weight $\alpha^i_{t'}$ for each transaction in the community to quantify the corresponding impact:
\begin{equation}
        \beta_{t'}={v_u\rm tanh}(W_{u_1} \tilde{x}_{t'}), \alpha^j_{t'}=\frac{{\rm exp}(\beta_{t'})}{\Sigma_{k \in C_{i}}{\rm exp}(\beta_{k})},
\end{equation}
where $W_{u_1} \in \mathbb{R}^{d_m \times d_{\tilde{x}}}, v_u \in \mathbb{R}^{1 \times d_m}$, and $C_j$ is the transaction set in community $j$.
The community embedding is computed by aggregating the environmental representation of each transaction with the attention weight:

\begin{equation}
    u^j_{t+1}={\rm ReLU}(W_{u_2}(\sum \limits_{t'\in C_j} \alpha^j_{t'}\tilde{x}_{t'})),
\end{equation}
where $W_{u_2} \in \mathbb{R}^{d_m \times d_{\tilde{x}}}$ is a learnable weight.

Then we propagated the embedding of the target transaction's community neighbors to the target transaction itself.
Instead of passing through the community of the target transaction, we focused on the connection between the target itself and the community neighbors to enable the flexibility for community modeling for the other target transactions in the same community.
Specifically, we calculated the attention weight for each community to engage with the influential community for the target transaction based on the concatenation of the target transaction's object embedding $o_{t+1}$ and the community embedding $u^j_{t+1}$: 
\begin{equation}
    \begin{split}
        \gamma^j_{t+1}&=v_c {\rm tanh}(W_{c_1} [o_{t+1} \oplus u^j_{t+1}]),\\
        \alpha^j_{t+1}&=\frac{{\rm exp}(\gamma^j_{t+1})}{\Sigma_{k \in N^c_{t+1}}{\rm exp}(\gamma^k_{t+1})},
    \end{split}
\end{equation}
where $W_{c_1} \in \mathbb{R}^{d_m \times 2d_m}$ and $v_c \in \mathbb{R}^{1 \times d_m}$ are learnable weights, and $N^c_{t+1}$ is the set of the target transaction's community neighbors.

Finally, the neighbor community embedding $c_{t+1}$ was aggregated by the community neighbors:
\begin{equation}
    c_{t+1}={\rm ReLU}(W_{c_2}(\sum \limits_{i \in N^c_{t+1}} \alpha^i_{t+1}u^i_{t+1})),
\end{equation}
where $W_{c_2} \in \mathbb{R}^{d_m \times d_m}$ is the learnable weight.

\subsection{Dynamic Adaptor}
Generally, prices of properties are influenced differently based on multiple factors.
For example, the demand of public transportation affects the real estate price in metropolitan areas, while this demand has fewer effects in remote areas since people mostly drive by themselves.
One of the potential solutions is to separate the prediction based on different districts.
However, this ignores the correlation between each region and is hard to learn for areas with insufficient transactions.
To this end, we propose the dynamic adaptor to adaptively learn the distribution of different target transactions to predict the real estate price.
 
Specifically, multiple learnable kernel weights are designed in the dynamic adaptor and are used to appraise the target transaction based on the input-generated weighted kernel.
To obtain the input-related importance of the kernels, we used the object embedding $o_{t+1}$, environment embedding $e_{t+1}$, neighbor relation embedding $r_{t+1}$ and neighbor community embedding $c_{t+1}$ of the target transaction to generate the target representation $h'_{t+1} \in \mathbb{R}^{4d_m}$
\begin{equation}
    h'_{t+1}=[o_{t+1} \oplus  e_{t+1} \oplus r_{t+1} \oplus c_{t+1}].
\end{equation}
Then we computed the attention weight of the $k$-th kernel weight:
\begin{equation}
        z^k_{t+1}=w_{k} \sigma(W_{r_1}h'_{t+1}), \pi^k_{t+1}=\frac{{\rm exp}(\frac{z^k_{t+1}}{\tau})}{\sum \limits^K_{l=1}{\rm exp}(\frac{z^l_{t+1}}{\tau})},
    \label{eq:dynamic_predictor_attention}
\end{equation}
where $W_{r_1} \in \mathbb{R}^{d_m \times 4d_m}, w_{k} \in \mathbb{R}^{1 \times d_m}$, and $K$ is the number of kernels.
It is noted that we also applied the softmax temperature to use near-uniform attention in early training epochs, which can better optimize multiple kernels simultaneously and avoid focusing on only one kernel before fully training the kernels.

To generate the input-related weights $\tilde{W}_{t+1} \in \mathbb{R}^{1 \times (4d_m+1)}, \tilde{b}_{t+1} \in \mathbb{R}^{1 \times 1}$ for the target transaction, we performed the aggregation of kernels based on these attention weights:
\begin{equation}
    \hat{W}_{t+1}=\sum \limits^K_{k=1}\pi^k_{t+1}\tilde{W}_k, \ \  \hat{b}_{t+1}=\sum \limits^K_{k=1}\pi^k_{t+1}\tilde{b}_k,
\end{equation}
where $\tilde{W}_k \in \mathbb{R}^{1 \times (4d_m+1)}, \tilde{b}_k \in \mathbb{R}^{1 \times 1}$ are learnable kernel weights of the $k$-th kernel.

Finally, we adopted linear regression with the weights of the aggregated kernel, and integrated the concatenation of the target representation $h'_{t+1}$, preliminary estimation $\tilde{p}_{t+1}$ as the final representation $h_{t+1}$ to predict the target appraisal value $\hat{p}_{t+1}$:

\begin{equation}
        h_{t+1}=[h'_{t+1} \oplus \tilde{p}_{t+1}], \hat{p}_{t+1}=\hat{W}_{t+1}h_{t+1}+\hat{b}_{t+1}.
\end{equation}

\noindent\textbf{Training Objective.} We minimized the mean square error loss to learn the prediction of real estate: 
\begin{equation}
    Loss=\frac{1}{|S|} \sum \limits_{s_{t+1}\in S}(\hat{p}_{t+1}-p_{t+1})^2,
\end{equation}
where $S$ is the set of target transactions and $p_{t+1}$ is the ground truth value of the target transaction.

\section{Experiments}
\label{sec:experiments}
In the experiment, we aim to study three research questions: \textbf{RQ1:} How does ReGram perform compared with different groups of models for real estate appraisal? \textbf{RQ2:} How does neighbor price affect prediction results? \textbf{RQ3:} How does each component of ReGram contribute to the effectiveness?

\begin{table}
    \begin{tabular}{c|ccc|c}
 \toprule
  & Train & Validation & Test & Total\\
 \midrule
 New Taipei & 89,964 & 4,349 & 2,474 & 96,787\\
 Taipei & 34,311 & 1,815 & 1,176 & 37,302\\
 Taoyuan & 61,592 & 2,074 & 1,170 & 64,836\\
 Taichung & 40,916 & 1,669 & 988 & 43,573\\
 Tainan & 17,710 & 586 & 423 & 18,719\\
 Kaohsiung & 38,335 & 1,875 & 1,142 & 41,352\\
 \midrule
 Total & 282,828 & 12,368 & 7,373 & 302,569\\
 \bottomrule
\end{tabular}
    \caption{The numbers of transactions.}
    \label{tab:building_stat}
\end{table}

\subsection{Experimental Setup}
\noindent\textbf{Dataset Description.}
Since there is no public real estate transaction dataset, the experiments were conducted on the collected dataset in Taiwan special municipalities, including New Taipei, Taipei, Taoyuan, Taichung, Tainan, and Kaohsiung\footnote{We note that both Beijing and Chengdu datasets used in \cite{DBLP:conf/kdd/0003LZZLD021} have not been released; thus, we could not test our model on them.}.
We collected the transaction data from Taiwan Real Estate Transaction Website\footnote{https://lvr.land.moi.gov.tw/} and the PoI data from E.Sun Bank.
The target transactions in the dataset ranged from 2015/7/1 to 2021/6/30, of which we used 3 months of transactions as validation data (2021/1 - 2021/3), 3 months of transactions as testing data (2021/4 - 2021/6), and the others as training data.
Table \ref{tab:building_stat} shows the statistics of the training, validation and testing transactions.

\noindent\textbf{Implementation Details.}
We performed standard normalization on numeric data, and used one-hot encoding for categorical data city by city.
For the object feature, the dimension $d_o$ ranged from 500 to 700, and the dimension of the environment feature $d_e$ was 1750.
The hyperparameters were tuned based on the validation set for each model.
The dimensions of both object embedding and environment embedding $d_m$ were set to 256, and we set the number of dynamic kernels $K$ to 8 and the number of attention heads $H$ to 8.
Besides, we took the softmax temperature $\tau=30$, and performed batch normalization \cite{ioffe2015batch} before the attention calculation in Equations \ref{eq:neighbor_attention} and \ref{eq:dynamic_predictor_attention}.
To learn the weight of our model, Adam optimizer \cite{kingma2014adam} was employed with a learning rate of 0.001 and a batch size of 64 for 50 epochs.
The hyperparameters were tuned based on the validation set and all the experiments were conducted on a GeForce RTX 2080 Ti 11GB GPU.
The best result in the comparison experiments is in boldface, while the second-best result is underlined.

\noindent\textbf{Baselines.}
To evaluate the performance of ReGram, we compared with the following models:
The \textbf{Machine Learning} based methods include (1) LR, (2) KNN, (3) SVR, and (4) LGBM. 
The \textbf{Deep Neural Network} based model (DNN) consists of multilayer perceptrons.
The \textbf{Graph-based Neural Network} (GNN) includes the following (1) The Graph Convolutional Network (GCN) aggregates the neighbors' features in equal weight, (2) the Graph Attention Network (GAT) aggregates the neighbors' features in the weight computed by node features, (3) MugRep is the state-of-the-art model in the real estate appraisal with a transaction module and a hierarchical community module in the multitask learning manner.
For fair comparisons, the embedding by our transaction projection without the price information was also used in DNN, GCN, GAT, and MugRep.
For the machine learning baselines, we report the testing score due to the non-stochasticity.
We trained the model in 5 different random seeds for the DNN and GNN models.
In the experiments, we report the average and standard deviation of these 5 testing scores.

\noindent\textbf{Evaluation Metrics.}
To evaluate the performance, we used Mean Absolute Percentage Error (MAPE) following \cite{DBLP:conf/kdd/0003LZZLD021}, which can be viewed as the prediction accuracy of the estimated value\footnote{We also tested the performance with 10\% and 20\% hit-rates to test the accuracy of the absolute percentage error less than 10\% and 20\% respectively in Appendix.}.

\subsection{Overall Performance}
\begin{table*}
    \small

\begin{tabular}{l|cccccc||c}
    \toprule
      Model/Municipalities  & New Taipei & Taipei & Taoyuan & Taichung & Tainan & Kaohsiung & Average \\
    \midrule
        LR & 98.24 & 412.06 & 14.17 & 15.75 & 18.98 & 15.57 & 95.80 \\
        SVR & 33.27 & 23.01 & 27.69 & 24.59 & 27.13 & 27.46 & 27.19 \\
        KNN & 8.67  & 10.23 & 10.04 & 13.14 & 15.65 & 11.99 & 11.62 \\
        LGBM & 8.37 & 9.48  & 10.46 & 13.16 & 16.12 & \textbf{10.70} & 11.38\\
    \midrule
        DNN & 8.03$\pm$0.20 & \underline{9.16$\pm$0.09} & 9.07$\pm$0.24 & \underline{11.38$\pm$0.09} & 17.13$\pm$0.16 & 11.34$\pm$0.15 & 11.02$\pm$0.16 \\
    \midrule
        GCN & 28.38$\pm$1.19 & 25.91$\pm$0.17 & 26.75$\pm$0.37 & 28.29$\pm$1.51 & 29.77$\pm$0.55 & 27.21$\pm$0.89 & 27.72$\pm$0.78 \\
        GCN + neighbor's price & 26.80$\pm$1.29 & 25.05$\pm$0.26 & 26.06$\pm$0.30 & 28.52$\pm$1.08 & 29.15$\pm$0.56 & 26.46$\pm$0.64 & 27.01$\pm$0.69 \\
        GAT & 10.42$\pm$0.09 & 12.07$\pm$0.06 & 12.68$\pm$0.25 & 13.42$\pm$0.17 & 18.18$\pm$0.40 & 12.28$\pm$0.11 & 13.18$\pm$0.18 \\
        GAT + neighbor's price & 9.71$\pm$0.09 & 11.56$\pm$0.18 & 11.95$\pm$0.15 & 13.20$\pm$0.14 & 16.11$\pm$0.32 & 12.52$\pm$0.10 & 12.51$\pm$0.16 \\
        MugRep & 10.86$\pm$0.26 & 11.70$\pm$0.15 & 11.25$\pm$0.28 & 13.20$\pm$0.33 & 16.83$\pm$1.17 & 12.23$\pm$0.43 & 12.68$\pm$0.44 \\
        MugRep + neighbor's price & \underline{7.97$\pm$0.14} & 9.45$\pm$0.56 & \underline{8.86$\pm$0.22} & 11.70$\pm$0.15 & \underline{14.52$\pm$0.21} & 11.79$\pm$0.37 & \underline{10.72$\pm$0.28} \\
    \midrule
        ReGram (ours) & \textbf{7.15±0.15} & \textbf{8.31±0.06} & \textbf{8.33±0.18} & \textbf{11.24±0.10} & \textbf{14.36±0.39} & \underline{10.92±0.12} & \textbf{10.05±0.17} \\
    \bottomrule
\end{tabular}
    \caption{Overall performance in Mean Absolute Percentage Error (MAPE). + neighbor's price is denoted as appraising the preliminary value of the target transaction by referencing neighbors' price information.}
    \label{tab:overall_building}
\end{table*}

\noindent\textbf{Performance Comparison.}
Table \ref{tab:overall_building} reports the experiment results of our model and baselines, which shows that our model consistently outperformed other baselines in terms of MAPE.
Specifically, ReGram reached at least 8.8\% of MAPE performance improvement compared to all the baselines, which also signifies the importance of the cooperation of ReGram modules and the price information from other transactions.

It can be seen that MugRep performed the best among the graph-based baselines since it utilizes the community-level information and original transaction feature.
However, the non-graph based methods achieved better performance compared to the graph-based methods in some municipalities.
This indicates that naively aggregating the transaction features also takes some noise into account and degrades the model performance, while our relation aggregator eliminates the aggregation issue by considering the relation between the target transaction and surrounding important neighbors.
Moreover, another reason for inferior performance of MugRep may be the lack of multi-source data in the dataset, which also shows the demand for a large variety of data for MugRep and the robustness of our proposed method.
It is worth noting that KNN achieves competitive performance by only averaging the price of nearest neighbors.
This demonstrates the importance and necessity of considering the price information of neighbors, which is also attributed in our model.

In order to further testify the effect of neighbors' prices, we aligned the settings of the graph-based baselines with ReGram by appraising the preliminary value of the target transaction by referencing neighbors' price information (denoted as + neighbor price).
It is observed that MugRep, GAT, and GCN reach 15\%, 5\%, and 3\% significant MAPE improvement after taking the price information of neighbors into account, which shows the effectiveness of using neighbors' prices and generalizability to be applied in any graph-based model.
ReGram still outperforms these models even when the neighbors' price information is considered in the baseline models, which illustrates the strength of our model's robustness.

\subsection{Ablation Study}

\begin{table}
    \small
    \begin{tabular}{>{\centering\arraybackslash}c >{\centering\arraybackslash}c >{\centering\arraybackslash}c >{\centering\arraybackslash}p{1.5cm}||>{\centering\arraybackslash}p{1.9cm}}
 \toprule
    \multicolumn{2}{c}{\makecell{Neighbor}} & 
    \multirow{2}{*}{Community} & 
    \multirow{2}{*}{Adaptor} & 
    \multirow{2}{*}{Average}  \\
    price & relation & & &\\
 \midrule
 \midrule
    && \checkmark & Dynamic &  10.54±0.21 \\
    &\checkmark& \checkmark & Dynamic &  10.12±0.15 \\
    & GAT & \checkmark & Dynamic & 10.68±0.27\\
    \checkmark&& \checkmark & Dynamic & 10.13±0.16 \\
    \checkmark&\checkmark&&Dynamic& 10.15±0.15\\
    \checkmark&\checkmark&\checkmark&Single& 10.15±0.11\\
    \checkmark&\checkmark&\checkmark&Region& 10.35±0.18\\
 \midrule
    \checkmark&\checkmark&\checkmark&Dynamic& \textbf{10.05±0.17}\\
 \bottomrule
\end{tabular}

    \caption{Ablation study of ReGram in MAPE.}
    \label{tab:ablation_building}
\end{table}

To better understand the contributions of each module, we replaced the neighbor aggregator with GAT using the same settings to generate relation embedding.
Furthermore, we not only changed our dynamic adaptor to a single adaptor, which uses one linear kernel weight to appraise the value for all the regions, but also applied region-aware multitask learning based on the district to learn independent adaptors.
Table \ref{tab:ablation_building} illustrates the results in terms of the average MAPE.
We summarize the observations as follows.

As expected, removing any one component from ReGram degrades the prediction performance.
In addition, it is also observed that adding the neighbor aggregator boosts the MAPE performance by 4.6\%, which demonstrates the effectiveness of our relation modeling.
Besides, the MAPE performance is significantly improved by 4\% if we only consider either the weighted relation embedding or the price information (the MAPE is improved from 10.54 to 10.12 and 10.13, respectively).
Furthermore, the MAPE performance results are inferior when applying GAT to model the relation embedding (the MAPE is degraded from 10.54 to 10.68 compared to ReGram without the neighbor aggregator), which also implies that directly aggregating the neighbors' features introduces some irrelevant noise (e.g., neighbors' house area, floor number) which hampers the performance, while our neighbor aggregator aggregates the relation between target transactions and their neighbors to alleviate this issue.

When we replaced a single fully connected layer (i.e., one kernel) in the final adaptor with a multi-adaptor based on region-aware separation as in \cite{DBLP:conf/kdd/0003LZZLD021}, the MAPE performance relatively deteriorated from 10.15 to 10.35, indicating that a region-aware adaptor is not suitable if the information is insufficient.
However, our dynamic adaptor with multiple kernels predicting the value of real estate based on the different transactions again demonstrates the robust capability.

\section{Conclusion}
\label{sec:conclusion}

In this paper, we present a neighbor relation graph learning framework, ReGram, for tackling the challenging real estate appraisal problem.
We constructed the community node from the transaction data and built the real estate network at the transaction level and community level to model the relation between the properties in various aspects.
To tackle the issue of directly aggregating the feature of the neighbors, we propose the neighbor aggregator to aggregate the information by modeling the relation between the target transaction and their neighborhoods, and predict the preliminary value of real estate.
Moreover, we introduce the dynamic adaptor by adaptively learning the distribution of different target transactions to consider the similarities and discrepancies between each target transaction.
Comprehensive experiments on the real-world dataset demonstrate the effectiveness of our model compared to state-of-the-art baselines in different scenarios.
For future research, we aim to explore the model explainability for real estate appraisal, which is an important topic in financial applications.

\section{Acknowledgments}
We would like to thank Fu-Chang Sun, Jefferey Lin, Hsien-Chin Chou, Chih-Chung Sung, and Leo Chyn from E.Sun bank for sharing data and discussing the findings. 

\appendix

\begin{appendices}


\begin{table*}
\begin{tabular}{>{\centering\arraybackslash}p{2.1cm}||>{\centering\arraybackslash}p{2cm}|>{\centering\arraybackslash}p{1.5cm}|>{\centering\arraybackslash}p{1.5cm}|>{\centering\arraybackslash}p{1.5cm}|>{\centering\arraybackslash}p{1.5cm}|>{\centering\arraybackslash}p{1cm}|>{\centering\arraybackslash}p{1cm}|>{\centering\arraybackslash}p{1cm}|>{\centering\arraybackslash}p{1cm}}
 \toprule
  & ReGram (ours) & MugRep & GAT & GCN & DNN & LGBM & KNN & SVR & LR\\
 \midrule
 New Taipei & \textbf{76.17±0.77} & 57.76±2.10 & 59.41±0.90 & 24.90±0.51 & \underline{72.13±0.99} & 68.51 & 68.63 & 18.55 & 57.68\\
 Taipei & \textbf{71.09±1.00} & 55.48±1.45 & 52.36±1.36 & 24.66±1.40 & \underline{66.17±0.77} & 61.48 & 61.90 & 24.91 & 48.13\\
 Taoyuan & \textbf{73.25±0.84} & 58.63±1.40 & 49.06±1.99 & 23.33±1.51 & \underline{68.56±0.79} & 57.09 & 63.16 & 24.44 & 47.35\\
 Taichung & \textbf{58.46±1.11} & 49.17±2.74 & 49.35±1.17 & 24.86±0.58 & \underline{57.59±0.73} & 46.86 & 49.80 & 26.82 & 38.26\\
 Tainan & \underline{43.78±1.04} & 37.26±5.54 & 32.15±2.15 & 19.67±1.25 & 36.97±0.61 & 36.41 & \textbf{43.97} & 21.28 & 31.91\\
 Kaohsiung & \textbf{59.67±0.69} & 54.19±1.73 & 52.47±0.69 & 24.34±0.75 & 57.74±0.80 & \underline{59.37} & 56.74 & 19.44 & 42.47\\
 \midrule\midrule
 Average & \textbf{63.74±0.91} & 52.08±2.49 & 49.13±1.38 & 23.63±1.00 & \underline{59.86±0.78} & 54.95 & 57.37 & 22.57 & 44.30\\
 \bottomrule
\end{tabular}

\caption{Overall performance in terms of a 10\% hit-rate.}
\label{tab:overall_building_hit10}
\end{table*}

\begin{table*}
\begin{tabular}{>{\centering\arraybackslash}p{2.1cm}||>{\centering\arraybackslash}p{2cm}|>{\centering\arraybackslash}p{1.5cm}|>{\centering\arraybackslash}p{1.5cm}|>{\centering\arraybackslash}p{1.5cm}|>{\centering\arraybackslash}p{1.5cm}|>{\centering\arraybackslash}p{1cm}|>{\centering\arraybackslash}p{1cm}|>{\centering\arraybackslash}p{1cm}|>{\centering\arraybackslash}p{1cm}}
 \toprule
  & ReGram (ours) & MugRep & GAT & GCN & DNN & LGBM & KNN & SVR & LR\\
 \midrule
 New Taipei & \textbf{94.62±0.76} & 86.27±0.86 & 88.70±0.37 & 45.55±1.69 & 93.03±0.41 & \underline{94.26} & 91.75 & 37.31 & 86.18\\
 Taipei & \textbf{93.10±0.21} & 85.48±0.93 & 84.61±0.57 & 48.52±0.58 & 92.16±0.18 & \underline{92.43} & 88.10 & 53.23 & 78.83\\
 Taoyuan & \textbf{92.60±0.60} & 85.35±0.58 & 80.92±0.48 & 45.68±0.98 & \underline{91.91±0.37} & 87.52 & 87.52 & 45.98 & 80.09\\
 Taichung & \textbf{86.03±0.63} & 80.83±1.36 & 78.62±0.35 & 47.11±1.62 & \underline{85.16±0.37} & 80.26 & 80.77 & 49.49 & 70.45\\
 Tainan & \textbf{75.32±0.89} & \underline{68.56±4.15} & 64.26±2.13 & 39.20±1.13 & 66.62±0.84 & 67.85 & 72.34 & 42.08 & 59.10\\
 Kaohsiung & \underline{85.66±0.81} & 82.84±1.71 & 81.42±0.73 & 46.58±0.60 & 83.63±0.98 & \textbf{86.34} & 81.26 & 44.31 & 73.12\\
 \midrule\midrule
 Average & \textbf{87.89±0.65} & 81.56±1.60 & 79.76±0.77 & 45.44±1.10 & \underline{85.42±0.53} & 84.78 & 83.62 & 45.40 & 74.63\\
 \bottomrule
\end{tabular}

\caption{Overall performance in terms of a 20\% hit-rate.}
\label{tab:overall_building_hit20}
\end{table*}

\section{Hyperparameters of Baselines}
\label{sec:app_hyperparameters}

In this section, we conduct a grid search on the validation data and list the selected hyperparameters of our baselines.

\noindent\textbf{LGBM.}
The shared hyperparameters of LGBM are listed as follows:
The boosting method is gbdt. The number of leaves is 31, the learning rate is 0.1, the number of estimators is 100, and both reg\_alpha and reg\_lambda are 0.
For the other different hyperparameters, we set max\_depth in Taoyuan as 11 and all the other data as -1. For min\_split\_gain, we set Taipei and Tainan as 0.05, New Taipei as 0.07, Taoyuan as 0.02, and all the others as 0.

\noindent\textbf{SVR.}
The kernel in SVR is rbf, and we set tol (tolerance for stopping criterion) as 0.001, C (a regularization parameter) as 2 and shrinking to true value. For the other hyperparameters, we set epsilon as 0.1 for all the data. And for hyperparameter gamma, we set New Taipei and Taoyuan as “scale”, and the other data as “auto”.

\noindent\textbf{KNN.}
The number of neighbors to calculate the price of the target transaction of New Taipei is 5, Taipei is 3, Taoyuan is 4, Taichung is 9, Tainan is 8, and Kaohsiung is 7.

\noindent\textbf{DNN and GNN.}
For DNN, GCN, GAT and MugRep, we set the dimension of object and environment embedding, learning rate, batch size and training epochs as the setting of ReGram. Additionally, for GAT, we set the number of heads as 8. For MugRep, we also set the dimension of all the other embedding in MugRep as 256.


\section{Implementation Environment}
In this paper, all the experiments were conducted on a machine with Intel Xeon Silver 4110 CPU 2.1GHz, GeForce RTX 2080 Ti 11GB GPU, and 252GB RAM in Ubuntu 18.04.

\section{Overall Performance with Hit-Rate}
\label{sec:other_metric_performance}
Table \ref{tab:overall_building_hit10} and Table \ref{tab:overall_building_hit20} report the overall performance of ReGram and our baselines with 10\% hit-rate and 20\% hit-rate.
It is also obvious that ReGram steadily outperforms other baselines in all the metrics on the average scores, which also demonstrates the effectiveness of our model.

\begin{figure}
  \centering
  \includegraphics[width=\linewidth]{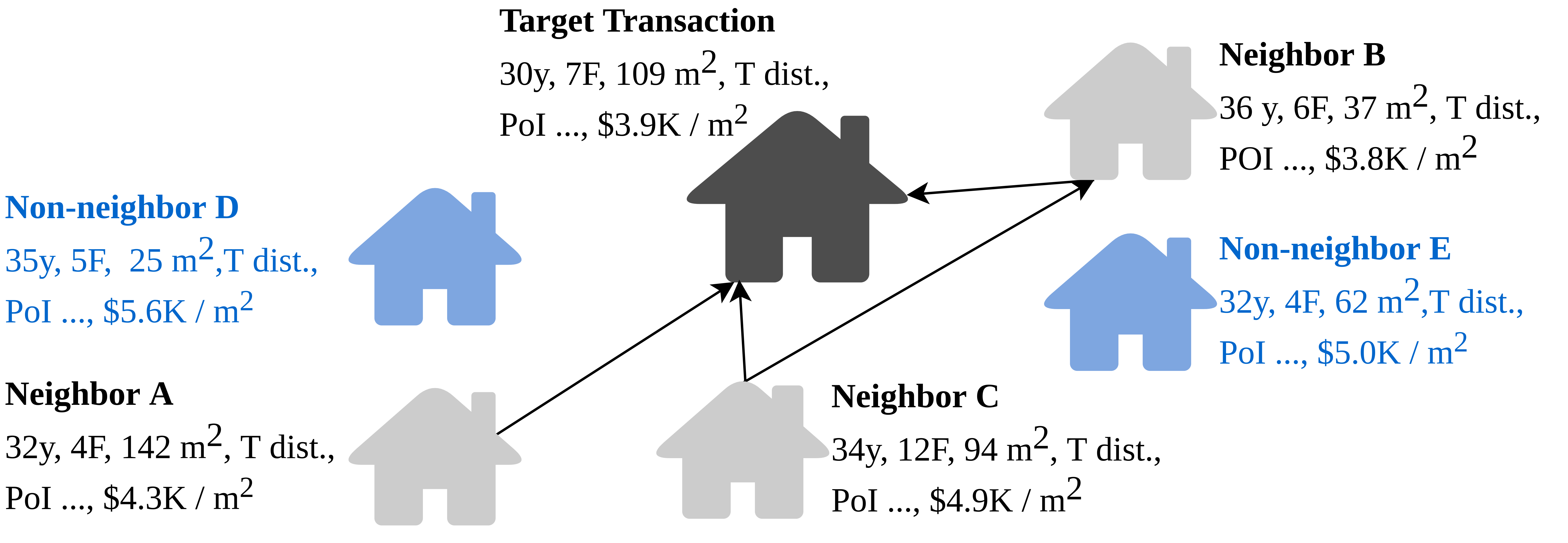}
  \caption{A scenario of the target transaction (black) and its neighbors (gray) and non-neighbors (blue) in Taipei.}
  \label{fig:ReGram_casestudy}
\end{figure}
 
\section{Case Study: Why Consider Neighbors?}

When executing a real estate transaction, buyers, sellers, and owners tend to consider information about the surroundings to investigate whether the target price is reasonable.
In general, characteristics of neighbors are one of the important factors to appraise real estate \cite{DBLP:conf/dsaa/NadaiL18}.
In this case, we aim to study the relations of geographic locations and surrounding information between the target transaction and its neighbors.
Figure \ref{fig:ReGram_casestudy} shows a target transaction and corresponding neighbors and non-neighbors in Taipei, which is the newest building compared with its neighbors.
The house areas of Neighbor A and Neighbor C are larger than that of the target transaction, and Neighbor B, Non-neighbor D, and Non-neighbor E are smaller.
All of the houses in this case are aged between 30 and 36 years.
Our ReGram appraises this target transaction with $\$ 4.0 K/m^{2}$, while the DNN model without considering neighbor information estimates $\$ 4.3 K/m^{2}$, which might be because its characteristics make it more likely to appraise a higher price.
However, it can be observed that neighbors' factors imply the situations in this area, \textit{e.g.}, the possible price range.
We note that although features of non-neighbors D and E are also similar to the target transaction but have higher prices, they are excluded due to the distance constraints of distances.
This case demonstrates a scenario of the reason to consider neighbor information to appraise real estate, and our model is capable of providing objective appraisal between triad relationships to bargain for the proper value of the property.


\end{appendices}

\bibliography{reference}

\end{document}